\documentclass[10pt,conference]{IEEEtran}
\IEEEoverridecommandlockouts

\usepackage{cite}
\usepackage{amsmath,amssymb,amsfonts}
\usepackage{algorithmic}
\usepackage{graphicx}
\usepackage{textcomp}
\usepackage{xcolor}
\def\BibTeX{{\rm B\kern-.05em{\sc i\kern-.025em b}\kern-.08em
    T\kern-.1667em\lower.7ex\hbox{E}\kern-.125emX}}
\usepackage{times}
\usepackage{latexsym}
\usepackage{amsmath}
\usepackage[T1]{fontenc}
\usepackage[utf8]{inputenc}
\usepackage{microtype}
\usepackage{inconsolata}
\usepackage{graphicx}
\usepackage{tabularx}
\usepackage{multirow}
\usepackage[utf8]{inputenc}
\usepackage{booktabs}
\usepackage{amsfonts}
\usepackage{enumitem}
\usepackage{url}
\usepackage{xspace} 
\usepackage[breaklinks]{hyperref}

\setenumerate[1]
{itemsep=0pt,partopsep=0pt,parsep=\parskip,topsep=5pt}
\setitemize[1]{itemsep=0pt,partopsep=0pt,parsep=\parskip,topsep=5pt}
\setdescription{itemsep=0pt,partopsep=0pt,parsep=\parskip,topsep=5pt}

\newcommand{\eg}{\textit{e.g.,}\xspace}
\newcommand{\ie}{\textit{i.e.,}\xspace}
\newcommand{\etc}{\textit{etc.}\xspace}
\newcommand{\etal}{\textit{et al.}\xspace}

\begin{document}

\title{{UniGenCoder}: Merging \textsc{Seq2Seq} and \textsc{Seq2Tree} Paradigms for Unified Code Generation}

\author{
	\IEEEauthorblockN{
		Liangying Shao\IEEEauthorrefmark{1}\textsuperscript{\textsection}, 
		Yanfu Yan\IEEEauthorrefmark{2}\textsuperscript{\textsection}, 
        Denys Poshyvanyk\IEEEauthorrefmark{2} and
		Jinsong Su\IEEEauthorrefmark{1}\textsuperscript{\pounds} } 
	\IEEEauthorblockA{\IEEEauthorrefmark{1}School of Informatics, Xiamen University, China}
	\IEEEauthorblockA{\IEEEauthorrefmark{2}William \& Mary, Virginia, USA}
	\IEEEauthorblockA{liangyingshao@stu.xmu.edu.cn, yyan09@wm.edu, denys@cs.wm.edu, jssu@xmu.edu.cn} 
    
} 

\maketitle
\begingroup\renewcommand\thefootnote{\textsection}
\footnotetext{Equal contribution.}
\endgroup
\begingroup\renewcommand\thefootnote{\pounds}
\footnotetext{Corresponding author.}
\endgroup

\begin{abstract}
Deep learning-based code generation has completely transformed the way developers write programs today. 
Existing approaches to code generation have focused either on the Sequence-to-Sequence paradigm, which generates target code as a sequence of tokens, or the Sequence-to-Tree paradigm, which outputs code as a sequence of actions. While these two paradigms are 
intuitively complementary, their combination has not been previously explored. By comparing the code generated under these two paradigms, we find that integrating them holds significant potential. In this paper, we propose UniGenCoder for code-related generation tasks, which consists of a shared encoder, a shared decoder with a minimal set of additional parameters to unify two paradigms, and a selector that dynamically chooses optimal paradigm for each instance. Also, during the model training, we first perform the multi-task learning and distillation strategies to facilitate knowledge transfer between two paradigms, and then leverage contrastive learning to train the selector. 
Experimental results on the text-to-code and code-to-code generation tasks demonstrate the effectiveness of our proposed model. We release our code at \url{https://github.com/DeepLearnXMU/UniGenCoder}.
\end{abstract}

\begin{IEEEkeywords}
Code Generation, Sequence-to-Sequence, Sequence-to-Tree
\end{IEEEkeywords}

\section{Introduction}
Code understanding and generation are fundamental tasks in software development, impacting a wide range of activities such as debugging, refactoring, program synthesis, \etc. As modern software systems grow increasingly complex, automating these tasks can significantly enhance developer productivity and streamline the overall software development lifecycle. Pre-trained Transformer-based \cite{DBLP:conf/nips/VaswaniSPUJGKP17} language models like BERT \cite{DBLP:conf/naacl/DevlinCLT19} and T5 \cite{DBLP:journals/jmlr/RaffelSRLNMZLL20} have achieved significant success in natural language (NL) understanding and generation. Given the similarities between NL and programming languages (PLs), it is common practice to adapt these models for PLs (such as CodeBERT \cite{DBLP:conf/emnlp/FengGTDFGS0LJZ20} and CodeT5 \cite{DBLP:conf/emnlp/0034WJH21}), treating code as a sequence of words to automate various SE tasks \cite{DL-survey,7180092,10.1145/2970276.2970326, athena}. Generally, pre-trained models are categorized into three types: encoder-only (\eg CodeBERT), decoder-only (\eg CodeGPT \cite{DBLP:conf/nips/LuGRHSBCDJTLZSZ21}), and encoder-decoder models (\eg CodeT5). Encoder-only models, optimized for understanding tasks, are less effective for generation, while decoder-only models, designed for generation, are suboptimal for understanding~\cite{DBLP:conf/emnlp/0034WJH21}. Encoder-decoder models can seamlessly support both code understanding and generation.

Despite the similarity between NL and PL, code is more strongly structured, with its semantics heavily relying on the structural dependency between code elements. As a result, researchers have started encoding structural information (\eg data flow~\cite{DBLP:conf/iclr/GuoRLFT0ZDSFTDC21} abstract syntax trees (ASTs) \cite{DBLP:conf/acl/GuoLDW0022}) into input code sequences for the encoder training, in order to learn structure-aware code representations through various pre-training tasks. On the decoder side, most approaches \cite{DBLP:conf/kbse/LiuLZJ20, DBLP:journals/corr/abs-2107-03374, DBLP:journals/corr/abs-2207-10397} output a continuous token sequence based on the Sequence-to-Sequence (Seq2Seq) paradigm, while the others \cite{DBLP:conf/acl/YinN17, DBLP:conf/aaai/SunZXSMZ20, DBLP:conf/emnlp/YinN18, DBLP:conf/acl/JiangZM00H0S20,  DBLP:conf/aaai/XieSGLCYW21} leverage syntactic information and decode code as an action sequence corresponding to the pre-order traversal of an AST, following the Sequence-to-Tree (Seq2Tree) paradigm. Both of these two paradigms mainly target code-related generation tasks, such as text-to-code and code-to-code generation, and code repair.

However, the combination of Seq2Seq and Seq2Tree paradigms has not yet been explored. Intuitively, Seq2Seq-based code models align with the paradigm used by NL models, enabling them to fully leverage the prior knowledge learned from large pre-train corpora. In contrast, while Seq2Tree-based code models can also be fine-tuned from NL models, their focus shifts toward capturing syntactic structure when generating ASTs. Despite considering the crucial syntax in code, this shift to a different paradigm may lead to the loss of some valuable knowledge from NL corpora. Therefore, combining two paradigms presents an opportunity to benefit from both. To verify our hypothesis, we compare the performance of CodeT5 under the Seq2Seq paradigm -- CodeT5(Seq2Seq), and under the Seq2Tree paradigm -- CodeT5(Seq2Tree)\footnote{The CodeT5 was originally designed under the Seq2Seq paradigm, so we implemented the CodeT5(Seq2Tree) ourselves to ensure a fair comparison.}, using BLEU scores on the CONCODE dev set, a common text-to-code generation dataset. Specifically, CodeT5 adapts the text-to-text transformer (\ie T5) model for code by incorporating token type information, and the BLEU score is widely-used to measure the similarity between the generated and ground-truth code. Our results show that even though CodeT5(Seq2Seq) is more effective than CodeT5(Seq2Tree) regarding the overall performance 
(37.06 vs. 35.57), a more fine-grained analysis reveals that CodeT5(Seq2Seq) outperforms CodeT5(Seq2Tree) on 34.95\% of instances, while CodeT5(Seq2Tree) excels in 40.00\% of instances. 
Additionally, CodeT5(Seq2Tree) predicts ground truth tokens with an average probability of 92.89\%, higher than the 80.98\% achieved by CodeT5(Seq2Seq). 

Therefore, in this paper, we propose a novel model, named UniGenCoder, designed to leverage the strengths of both Seq2Seq and Seq2Tree paradigms. Built upon the Transformer-based CodeT5 model, UniGenCoder features an \textit{improved decoder} that unifies token sequence generation and action sequence (\ie AST) generation, along with an additional \textit{dynamic selector} that chooses the appropriate paradigm for each input instance (\ie whether to generate a token sequence or an action sequence). We further introduce a two-stage strategy for the model training. Specifically, we first adopt a multi-task learning objective to simultaneously learn both token and action sequence generations, and use a distillation objective to enforce knowledge transfer between the Seq2Seq and Seq2Tree paradigms. In the second stage, we leverage contrastive learning to tune the selector, with other model parameters fixed.

As our primary technical contribution is mainly on the decoder side, we evaluate our UniGenCoder on two code-related generation tasks: text-to-code and code-to-code generation. The experimental results demonstrate the effectiveness of our model architecture and its training strategy when compared to other state-of-the-art \textit{medium-scale} large language models (LLMs) for code. Notably, our proposed framework and training strategy are backbone-agnostic, making them directly applicable to any Transformer-based LLMs, particularly improving decoder-only and encoder-decoder models. Moreover, the unification of Seq2Seq and Seq2Tree paradigms within a single model remains underexplored by \textit{large-scale} LLMs as well. Therefore, in future work, we plan to extend our work to large-scale LLMs (\eg, Code Llama \cite{DBLP:journals/corr/abs-2308-12950}, StarCoder~\cite{li2023starcodersourceyou}) to achieve superior performance as long as they are open-source.

\section{Related Work}
Most code generation models adopt the Seq2Seq paradigm to output a continuous sequence of tokens. Recognizing the importance of structural information in code, recent approaches have begun integrating this information into input sequences to enhance code representations. For example, GraphCodeBERT~\cite{DBLP:conf/iclr/GuoRLFT0ZDSFTDC21} first incorporates code structure (\ie data flow graphs extracted from ASTs) into input during pre-training. 
CodeT5~\cite{DBLP:conf/emnlp/0034WJH21} leverages node type information from ASTs and introduces an identifier-aware pre-training objective. UniXcoder~\cite{DBLP:conf/acl/GuoLDW0022} incorporates ASTs through a one-to-one mapping method that transforms an AST into a sequence.

In contrast to the aforementioned pre-trained models that integrate ASTs into the input, some models~\cite{DBLP:conf/aaai/XieSGLCYW21, DBLP:conf/acl/JiangZM00H0S20} focus on learning AST structures at the decoding stage. These models adopt the Seq2Tree paradigm, generating code as the corresponding pre-order traversal of ASTs, allowing them to better capture code syntax during generation. For instance, Xie \etal ~\cite{DBLP:conf/aaai/XieSGLCYW21} leverage mutual distillation across various AST traversal strategies to enhance action prediction accuracy. Jiang \etal~\cite{DBLP:conf/acl/JiangZM00H0S20} improve the Seq2Tree paradigm by designing a context-based Branch Selector to determine optimal expansion orders for multi-branch nodes. Note that existing Seq2Tree approaches are trained from scratch (\ie they are not trained/fine-tuned from pre-trained code models).


Unlike previous works, our paper proposes a novel model architecture that allows the model to effectively master both paradigms, enhancing its generative capabilities through a multi-task learning strategy, a distillation strategy, and a paradigm selector. To the best of our knowledge, this is the first work to unify the Seq2Seq and Seq2Tree paradigms in a single model.
Recently, we notice the similarity between \cite{10.1145/3609437.3609465} and our study, which also realizes the complementary advantages of the two paradigms through distillation loss. However, our work further enhances the efficiency of the fusion of the two paradigms by introducing a pre-trained model, designing a shared decoder, and developing a contrastive learning selector.

\begin{figure}[t]
\centering
\footnotesize
\includegraphics[width=0.51\textwidth, trim=104 135 80 145, clip]{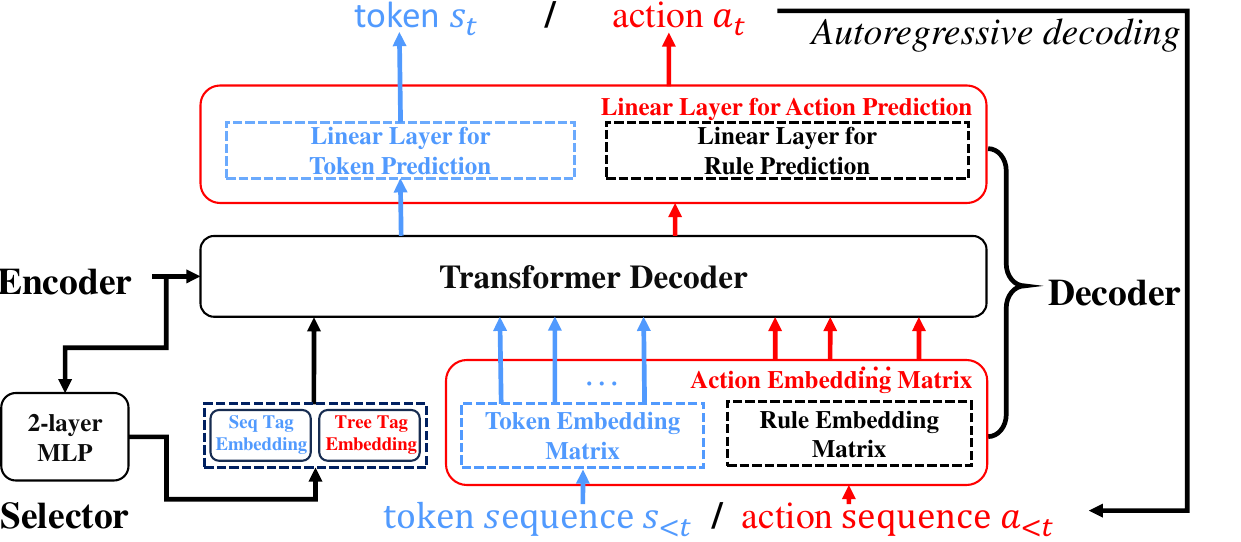}
\setlength{\abovecaptionskip}{0pt}
\caption{The overall architecture of our UniGenCoder model. The token embedding matrix and linear layer for token prediction are designed for the Seq2Seq paradigm, while the action embedding matrix and linear layer for action prediction are tailored to the Seq2Tree paradigm. Note that the token embedding matrix is included in the action embedding matrix, as output actions in the Seq2Tree paradigm can be either tokens or rules. Likewise, the linear layer for token prediction is contained by the linear layer for action prediction.} 
\label{fig:overview}
\vspace{-0.4cm}
\end{figure}

\section{Our Model}

In this section, we first introduce the basic architecture of our model, followed by a detailed explanation of the two-stage training strategy we designed for its optimization.

\subsection{Model architecture}
As shown in Figure \ref{fig:overview}, our model consists of a backbone encoder-decoder network (\ie CodeT5) shared by two paradigms, with an additional selector. We utilize the vanilla Transformer encoder and enhance the Transformer decoder by unifying token (\ie Seq2Seq) and action sequence generation (\ie Seq2Tree). The added selector dynamically chooses the appropriate paradigm for each input instance. Below, we describe the decoder and the selector in detail.

\textbf{Decoder.} This module extends the Transformer decoder with minimal additional parameters, including two tag embeddings, rule embeddings, and a corresponding linear layer that converts the decoder output into predicted rule probabilities.
According to Figure~\ref{fig:overview}, our decoder receives a specific tag to indicate the optimal generation paradigm for each instance. Specifically, the \textit{seq} tag would instruct our decoder to generate a token at each timestep using the token embedding matrix and the linear layer for token prediction. The \textit{tree} tag would direct our decoder to generate an action, either a token or a rule, at each timestep based on the action embedding matrix and the linear layer for action prediction. Note that the token embedding matrix and linear layer for token prediction are also included within the action embedding matrix and linear layer for action prediction.

\textbf{Selector.} This module is a two-layer MLP with ReLU activation. By feeding the average of the encoder’s last hidden states, $h_e$, into the selector, we obtain the selection probability distribution $p_s$ of the instance as $\mathrm{softmax}( W_2\cdot\mathrm{ReLU} (W_1h_e  + b_1 ) + b_2), $ where $W_*$ and $b_*$ are corresponding to the parameter matrices and bias terms for each layer, respectively. According to the tag with the higher selection probability, the decoder then outputs code under the paradigm indicated by this tag.

\begin{figure*}[t]
\vspace{-0.8cm} 
\centering
\footnotesize
\includegraphics[width=0.83\textwidth, trim=70 175 78 150, clip]{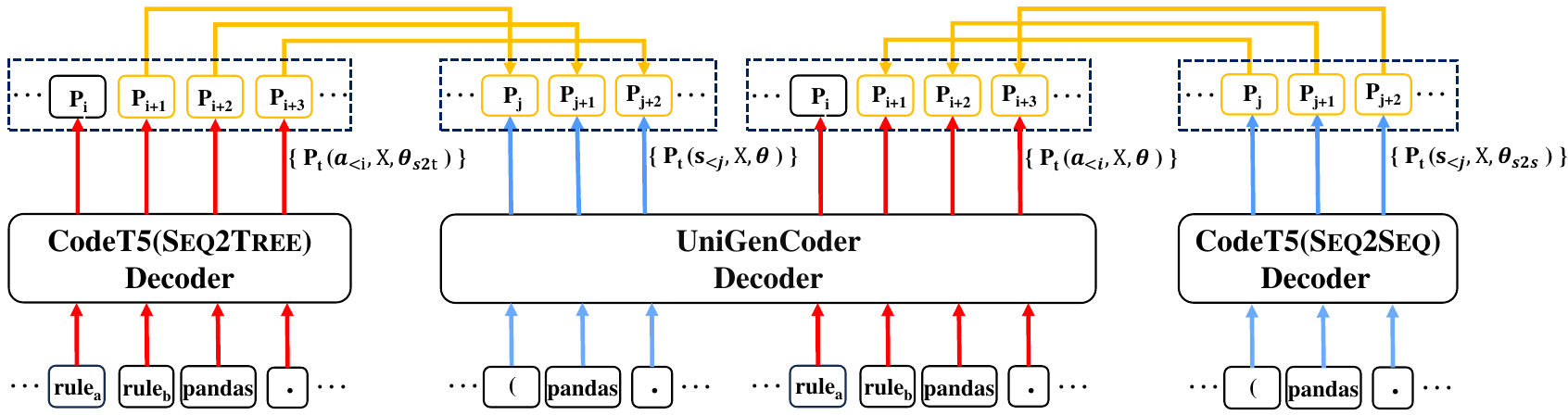}
\caption{Our proposed distillation strategy. $\theta, \theta_{s2s}, \theta_{s2t}$ represent the parameters of UniGenCoder, CodeT5(Seq2Seq) and CodeT5(Seq2Tree), respectively.}
\label{fig:distill}
\vspace{-0.3cm}
\end{figure*}

\subsection{Training Objectives}
We fine-tune our UniGenCoder from the existing CodeT5-base model. Specifically, we first use multi-task learning and knowledge distillation to train our backbone network, and then adopt contrastive learning to tune the selector with backbone parameters fixed.

\subsubsection{Backbone Training} 
given $(X, (s_1, ..., s_{n}), (a_1, ..., a_{m}))$, where $X$, $s_t$, $a_t$ are the input sequence, the $t$-th token, the $t$-th action, repsectively, 
we first train our backbone network with the following objective:
\begin{equation}
\setlength{\abovedisplayskip}{3pt}
\label{model_loss}
 \mathcal{L} = (1 - \lambda)\cdot  \mathcal{L}_{m} + \lambda \cdot  \mathcal{L}_{kd},
\setlength{\belowdisplayskip}{3pt}
\end{equation}
\setlength{\belowdisplayskip}{3pt}

\noindent where $\lambda$ is a hyperparameter used to balance effect of the multi-task learning loss  $\mathcal{L}_{m}$ and the distillation loss  $\mathcal{L}_{kd}$.

\textbf{Multi-task Learning Loss $\mathcal{L}_{m}$.} Via this 
loss, we train the model to jointly learn the generations of both token sequence and action sequence. Formally, 
$\mathcal{L}_{m}$ is defined as follows:
\setlength\abovedisplayskip{3pt}
\begin{align}
     \mathcal{L}_{m} &=  \mathcal{L}_{s} + \mathcal{L}_{t} \nonumber \\
     &= -(\frac{1}{n}\sum_{j=1}^{n}{\log\, p(s_j | s_{<j}, X; \theta)} \nonumber \\
     & \quad + \frac{1}{m}\sum_{i=1}^{m}{\log\, p(a_i | a_{<i}, X; \theta)} ), 
\end{align}
\setlength\belowdisplayskip{3pt}

\noindent where $s_{<i}$ represents the tokens appeared before $s_i$, $a_{<j}$ represents the actions appeared before $a_j$, and $\theta$ denotes the parameters of the backbone network. $\mathcal{L}_{s}$ and $\mathcal{L}_{t}$ are the cross-entropy losses for the generations of token sequence and action sequence, respectively. 

\textbf{Distillation Loss $\mathcal{L}_{kd}$.} To facilitate knowledge transfer between two paradigms, we incorporate a distillation strategy into the backbone training process. As illustrated in Figure \ref{fig:distill}, we first fine-tune CodeT5(Seq2Tree) and CodeT5(Seq2Seq) as teacher models using the same dataset intended for UniGenCoder training. We then simultaneously transfer their knowledge to our model through the distillation loss $\mathcal{L}_{kd}$:
\begin{align}
\setlength{\abovedisplayskip}{3pt}
\mathcal{L}_{kd} &= \mathrm{KL}(P(s_{<j}, X; \theta) || P(a_{<i}, X; \theta_{s2t})) \   + \\ \nonumber
&\quad\  \mathrm{KL}(P(a_{<i}, X; \theta) || P(s_{<j}, X; \theta_{s2s})), 
\setlength{\belowdisplayskip}{3pt}
\end{align}
where $\mathrm{KL}(\cdot|\cdot)$ represents the Kullback-Leibler divergence between two distributions, and $\theta_{s2s}$ and $\theta_{s2t}$ represent the model parameters of CodeT5(Seq2Seq) and CodeT5(Seq2Tree), respectively. The $j$-th ground truth token corresponds to the $i$-th ground truth action. Specifically, since the action sequence consists of both rules and tokens, each token in the target token sequence has a corresponding action in the target action sequence. During the distillation process, we focus solely on these corresponding actions and draw closer the predicted probability distributions for each target token with its corresponding action.

\subsubsection{Selector Training}
we then fix all backbone parameters and leverage contrastive learning to tune the selector.
To this end, we use the backbone to generate a token sequence and an action sequence for each input instance. If the BLEU score of the token sequence is greater than that of the action sequence, we assign the instance with the \textit{seq} tag, and vice versa. We discriminately train the selector with a max-margin loss:
\begin{equation}
\setlength{\abovedisplayskip}{3pt}
 \mathcal{L}_{c} = \sum \max\{0, 1 - (p_{s}^{+} - p_{s}^{-})\},
\setlength{\belowdisplayskip}{3pt}
\end{equation}
where $p_{s}^{+}$ is the probability of the selector with the assigned tag, 
and $ p_{s}^{-}$ denotes that of the selector with the other tag.

\section{Experiment Design} 
To evaluate UniGenCoder's performance in text-to-code and code-to-code generation tasks, we investigate the following research questions (RQs):

\begin{enumerate}[label=\textbf{RQ$_\arabic*$:}, ref=\textbf{RQ$_\arabic*$}, wide, labelindent=5pt]\setlength{\itemsep}{0.2em}
    \item {\textit{How effective is UniGenCoder as compared to state-of-the-art medium-scale code-based LLM baselines?}}
    \item {\textit{How do different modules in UniGenCoder impact its overall performance? }}
\end{enumerate}

\vspace{-0.2cm} 

\subsection{Evaluation Tasks and Datasets}
We consider two generation tasks\footnote{We leverage tree-sitter to convert all codes into AST. \url{https://github.com/tree-sitter/tree-sitter-java}}: text-to-code and code-to-code generation. Text-to-code generation was carried out on the \emph{CONCODE} dataset \cite{DBLP:conf/emnlp/IyerKCZ18}, which involves generating Java functions given NL descriptions and class environment contexts. \emph{CONCODE} contains 100K, 2K and 2K examples for training, validation and test, respectively, which is widely-used by the popular \emph{CodeXGLUE}~\cite{DBLP:conf/nips/LuGRHSBCDJTLZSZ21} benchmark and a number of pre-trained code models \cite{DBLP:conf/naacl/AhmadCRC21, DBLP:conf/nips/LuGRHSBCDJTLZSZ21, DBLP:conf/acl/GuoLDW0022, DBLP:conf/emnlp/0034WJH21}. As for code-to-code generation (\ie code translation), we currently focus on translating C\# code to Java with equivalent functionality, utilizing the dataset provided by the \emph{CodeXGLUE} benchmark. We have not conducted experiments to translate Java code to C\# due to additional effort in extracting syntactic rules for C\#, but plan to include this in future work. Note that we can directly reuse the extraction rules for Java that have been developed for the text-to-code generation task. This dataset consists of paired (C\#, Java) functions with 10,300, 500 and 1,000 samples for training, validation and test, respectively.

\vspace{-0.1cm} 
\subsection{Evaluation Details}

\textbf{Evaluation Metrics.} Following CodeT5 \cite{DBLP:conf/emnlp/0034WJH21}, we adopt standard metrics for evaluation, \ie exact match (EM), BLEU~\cite{lin2004orange}, and CodeBLEU~\cite{DBLP:journals/corr/abs-2009-10297} for text-to-code generation, and EM and BLEU for code-to-code generation. EM measures whether the generated code exactly matches the ground-truth code. BLEU evaluates n-gram (specifically, 4-gram) overlap between the generated and reference code. CodeBLEU extends BLEU by also considering syntactic and semantic overlap.

\textbf{Baselines.} We compare UniGenCoder against two categories of  state-of-the-art code models (all of similar size (100-200M) to UniGenCoder):  (i) models that do not leverage AST information during pre-training, including PLBART~\cite{DBLP:conf/naacl/AhmadCRC21}, CodeGPT~\cite{DBLP:conf/nips/LuGRHSBCDJTLZSZ21} and CodeGPT-adapted~\cite{DBLP:conf/nips/LuGRHSBCDJTLZSZ21}. (ii) models that do incorporate AST information during pre-training, including UniXcoder \cite{DBLP:conf/acl/GuoLDW0022} and CodeT5 (\ie CodeT5-base)~\cite{DBLP:conf/emnlp/0034WJH21}. Since existing Seq2Tree approaches for code generation are not based on pre-trained Transformer-based code models, 
we implemented the classical Seq2Tree paradigm \cite{DBLP:conf/emnlp/YinN18} using the pre-trained CodeT5 model as the backbone (\ie CodeT5(Seq2Tree)) for comparison. In the future, we plan to adapt other state-of-the-art Seq2Tree methods to pre-trained code models for further validation. 

\textbf{Hyperparameters.} We train UniGenCoder on an NVIDIA RTX 3090 GPU. The hyperparameters for training UniGenCoder’s encoder and decoder are consistent with those used by CodeT5. For training the selector, we use the Adam optimizer~\cite{DBLP:journals/corr/KingmaB14} with early stopping, a batch size of 16, and a peak learning rate of 1e-4 with linear decay. The selector's layer sizes are set to 512 and 2, respectively. $\lambda$ controls the balance between the multi-task learning loss and the distillation loss. We experimented with $\lambda$ values ranging from 0.1 to 0.9 in increments of 0.2, ultimately selecting $\lambda = 0.1$, which yielded the best performance.

\section{Results and Analysis}

\subsection{RQ$_1$: Main Results}

Table~\ref{table:results} presents the effectiveness of UniGenCoder compared to other medium-scale LLM baselines on text-to-code and code-to-code generation tasks. For text-to-code generation, our model outperforms the strongest baseline, CodeT5(Seq2Seq), by 4.10\%, 3.21\%, and 3.43\% ((44.36-42.89)/42.89) on EM, BLEU, and CodeBLEU metrics, respectively. While UniXcoder$\dag$ achieves similar performance to CodeT5(Seq2Seq)$\dag$ on EM, it falls short in terms of BLEU, making CodeT5(Seq2Seq)$\dag$ the best-performing baseline. We use the reproduced results of CodeT5(Seq2Seq) in our hardware environment to make a fair comparison with UniGenCoder. Additionally, although UniGenCoder is of same size (\# parameters) as CodeT5(Seq2Tree), the latter performs worse than CodeT5(Seq2Seq) and significantly below our model, indicating that UniGenCoder's superior performance is not due to a slight increase of model parameters. Beyond the effectiveness, we also measure the inference time of CodeT5(Seq2Seq) , CodeT5(Seq2Tree)  and our model. The results show that CodeT5(Seq2Tree) is three times slower than CodeT5(Seq2Seq) due to the action sequence being approximately three times longer than token sequence. However, our model is 2.3 times faster than CodeT5(Seq2Tree) because of the appropriate selection between two paradigms.

As for code-to-code generation, our UniGenCoder improves the best-performing baseline CodeT5(Seq2Seq) by 2.26\% and 4.97\% on EM and BLEU metrics, respectively. Currently, we only compare our model with baselines whose performance on these tasks has been evaluated/reported in their original papers, thus leaving some cells in Table~\ref{table:results} blank.

\begin{table}[t]
\caption{Experimental results on text-to-code and code-to-code generation tasks. \textbf{``$\dag$''} denotes the results are cited from the corresponding paper. Rows without the $\dag$ symbol are our reproduction or the results of our model.  \textbf{``-''} signifies that the corresponding paper did not evaluate their model on that task. \textbf{``Params''} represents the number of model parameters. }
\vspace{-0.2cm}
\setlength{\tabcolsep}{0.8mm}
\footnotesize
\centering
\begin{tabular}{l|c|cc|cc}
\toprule
\multicolumn{1}{c|}{\multirow{2}{*}{\textbf{Model}}} & \multicolumn{1}{c|}{\multirow{2}{*}{\textbf{ Params}}} &
\multicolumn{2}{c|}{{\bf Generation}} &
\multicolumn{2}{c}{{\bf Translation}} \cr
\multicolumn{1}{c|}{} & \multicolumn{1}{c|}{}& \textbf{EM} & \textbf{ BLEU/CodeBLEU} & \textbf{ EM} & \textbf{ BLEU} \\
\midrule
\multicolumn{6}{c}{\textit{Baseline without the AST Information}}\cr
\midrule
\multirow{1}{*}{PLBART$\dag$} & 140M
& 18.75 & 36.69 / 38.52 & 78.35 & 65.00\\
\multirow{1}{*}{CodeGPT$\dag$} & 124M
& 18.25 & 28.69 / 32.71 & - & -\\
\multirow{1}{*}{CodeGPT-adapted$\dag$} & 124M
& 20.10 & 32.79 / 35.98 & - & - \\
\midrule
\multicolumn{6}{c}{\textit{Baseline using the AST Information}}\cr
\midrule
\multirow{1}{*}{UniXcoder$\dag$} & 124M
& 22.60 & 38.23 / - & - & - \\
\multirow{1}{*}{CodeT5(Seq2Seq)$\dag$}&220M
& 22.30 & 40.80 / 43.20 & 79.87 & 66.90 \\
\multirow{1}{*}{CodeT5(Seq2Seq)} & 220M
& 21.95 & 40.46 / 42.89 & 83.79 & 68.40\\
\multirow{1}{*}{CodeT5(Seq2Tree)} & 225M
& 21.35 & 39.48 / 42.66 & 80.03 & 64.70 \\
\midrule
\multicolumn{6}{c}{\textit{Our model}}\cr
\midrule
\multirow{1}{*}{UniGenCoder} & 225M
& \textbf{22.85} & \textbf{41.76} / \textbf{44.36} & \bf 85.69 & \bf 71.80\\
\multirow{1}{*}{\quad \textit{w/o Selector}} & 224M
& 22.27 & 40.54 / 43.39 & 84.15 & 68.90\\
\multirow{1}{*}{\quad \textit{w/o Seq2Tree}} & 220M
& 22.65 & 41.06 / 43.73 & 85.23 & 70.90\\
\multirow{1}{*}{\quad \textit{w/o Seq2Seq}} & 224M
& 22.05 & 39.81 / 42.66 & 84.33 & 70.50\\
\bottomrule
\end{tabular}
\label{table:results}
\vspace{-0.6cm} 
\end{table}


\vspace{-0.1cm} 
\subsection{RQ$_2$: Ablation Study}
We report the performance of three variants of our model: (i) \textit{\textbf{w/o Selector}}, which removes the dynamic paradigm selector and randomly selects the paradigm, (ii) \textit{\textbf{w/o Seq2Tree}}, where the decoder is only based on the Seq2Seq paradigm but enhanced by CodeT5(Seq2Tree) via knowledge distillation. (iii) \textit{\textbf{w/o Seq2Seq}}, where decoder is only based on the Seq2Tree paradigm but enhanced by CodeT5(Seq2Seq) via knowledge distillation. As shown in Table~\ref{table:results}, removing the selector leads to a significant drop in performance, highlighting the benefit of dynamic paradigm selection. Moreover, both UniGenCoder \textit{w/o Seq2Seq} and \textit{w/o Seq2Tree} is less effective than our full model, but still outperforms CodeT5(Seq2Tree) and CodeT5(Seq2Seq), respectively. This demonstrates the necessity of each paradigm and the effectiveness of our distillation strategy which facilitates knowledge transfer between the two paradigms.

\section{Conclusion and Future Plan}
In this paper, we introduce UniGenCoder, a unified model that combines the Seq2Seq and Seq2Tree paradigms. Our experiments on text-to-code and code-to-code generation tasks demonstrate the effectiveness of the proposed model framework and training strategy. For future work, we plan to apply our framework to larger-scale code LLMs(\eg, Code Llama \cite{DBLP:journals/corr/abs-2308-12950}, StarCoder~\cite{li2023starcodersourceyou}), given that both medium- and large-scale LLMs are Transformer-based. Our framework is designed to benefit both decoder-only and encoder-decoder LLMs. Additionally, we intend to evaluate our model on more benchmarks (\eg HumanEval-X \cite{zheng2023codegeex}, MultiPL-MBPP \cite{DBLP:journals/corr/abs-2108-07732}) recently constructed for evaluating large-scale LLMs on text-to-code generation across multiple PLs. We also aim to explore other code-related generation tasks, such as code repair \cite{8668043, 8827954}. 

\section{Acknowledgment}
The project was supported by 
National Natural Science Foundation of China (No. 62276219),  
Natural Science Foundation of Fujian Province of China (No. 2024J011001),
and
the Public Technology Service Platform Project of Xiamen (No. 3502Z20231043).
We also thank the reviewers for their insightful comments.



\bibliographystyle{IEEEtran}
\bibliography{mybibfile.bib}

\end{document}